\documentclass[10pt,twocolumn,letterpaper]{article}

\usepackage[pagenumbers]{cvpr} 
\usepackage{algorithm}
\usepackage{algorithmic}
\usepackage{amsmath}
\usepackage{amsfonts}
\usepackage{textcomp}
\usepackage{todonotes}
\usepackage{booktabs} 
\usepackage{bbding}
\usepackage{pifont}

\usepackage{makecell}
\usepackage{booktabs} 
\usepackage{amsthm}
\newtheorem{definition}{Definition}
\usepackage{appendix}
\usepackage{tabularx}

\usepackage{times}  
\usepackage{helvet}  
\usepackage{courier}  
\usepackage[hyphens]{url}  
\usepackage{graphicx} 
\usepackage{natbib}  
\usepackage{caption} 

\usepackage{newfloat}
\usepackage{listings}
\usepackage{amsthm}    
\usepackage{amssymb}   

\definecolor{cvprblue}{rgb}{0.21,0.49,0.74}
\usepackage[pagebackref,breaklinks,colorlinks,allcolors=cvprblue]{hyperref}

\def\confName{CVPR}
\def\confYear{2026}

\title{Test-Time Adaptation for Tactile-Vision-Language Models}

\author{%
Chuyang Ye$^{1,2}$\textsuperscript{\textdagger}\thanks{Work completed at Shenzhen Technology University.},
Haoxian Jing$^{3}$\textsuperscript{\textdagger}, Qinting Jiang$^{4}$, Yixi Lin$^{1}$, Qiang Li$^{1}$,
Xing Tang$^{1}$, Jingyan Jiang$^{1}$\\
$^{1}$Shenzhen Technology University, $^{2}$New York University,$^{3}$ Shenzhen University, $^{4}$Tsinghua University \\
{\tt\small chuyang.ye@nyu.edu, jqt23@mails.tsinghua.edu.cn,
\{tangxing, jiangjingyan, liqiang\}@sztu.edu.cn}\\
{\small \textsuperscript{\textdagger}Equal contribution.}%
}

\providecommand{\thetitle}{Test-Time Adaptation for Tactile-Vision-Language Models}

\begin{document}
\maketitle
\begin{abstract}
Tactile-vision-language (TVL) models are increasingly deployed in real-world robotic and multimodal perception tasks, where test-time distribution shifts are unavoidable. Existing test-time adaptation (TTA) methods provide filtering in unimodal settings but lack explicit treatment of modality-wise reliability under asynchronous cross-modal shifts, leaving them brittle when some modalities become unreliable. We study TTA for TVL models under such shifts and propose a reliability-aware framework that estimates per-modality reliability from prediction uncertainty and perturbation-based responses. This shared reliability signal is used to (i) filter unreliable test samples, (ii) adaptively fuse tactile, visual, and language features, and (iii) regularize test-time optimization with a reliability-guided objective. On the TAG-C benchmark and additional TVL scenarios, our approach consistently outperforms strong TTA baselines, achieving accuracy gains of up to 49.9\% under severe modality corruptions, underscoring the importance of explicit modality-wise reliability modeling for robust test-time adaptation.
\end{abstract}
\section{Introduction}
 The integration of tactile sensing with vision and language, forming Tactile-Vision-Language (TVL) models, heralds a new frontier in robotics, promising to endow machines with a more holistic and human-like understanding of the physical world~\cite{li2023see}. This multi-modal synergy is pivotal for complex manipulation and perception tasks, such as object grasping and interaction~\cite{fenganytouch}. However, despite their potential, existing research on TVL models has largely overlooked a critical challenge in real-world deployment: the presence of distribution shifts at test time. In unconstrained environments, the data encountered by robots can differ dramatically from the curated distributions seen during training, posing a substantial obstacle to robust and reliable multi-modal perception.
\begin{figure}[t]
    \centering
    \includegraphics[width   =\linewidth]{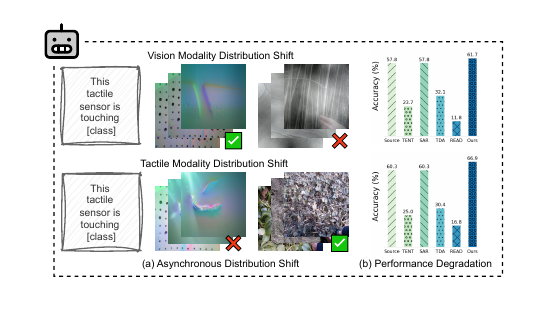}
    \caption{\textbf{Illustration of Asynchronous Distribution Shifts} in the Wild Multi-Modal Settings. \textbf{(a)} Illustration of asynchronous distribution shifts in vision and tactile modalities, where one modality experiences corruption (\ding{55}) while the other remains reliable (\checkmark), simulating unpredictable real-world conditions such as sensor degradation during a robot's rescue mission in disaster zones. \textbf{(b)} Performance degradation under these shifts, comparing Clip-based Zero-shot classification accuracy (\%) of baseline methods (Source, TENT, SAR, READ) against our RobustTouch framework, highlighting its superior adaptation to maintain robust perception amid modality-specific uncertainties.}
    \label{fig:motivation1}
\end{figure}

 Test-time adaptation (TTA) has recently emerged as an effective strategy for enhancing model resilience to such distribution shifts~\cite{wangtent,zhang2022memo,niutowards,lee2024entropy}. By enabling models to adapt online to unlabeled test data, TTA has shown considerable success in unimodal settings, leveraging self-supervised objectives and entropy minimization to maintain performance under changing conditions.  Recent multi-modal TTA approaches have extended these concepts to scenarios involving multiple modalities~\cite{yang2024test,zhao2025attention,guosmoothing}. These approaches typically attempt to achieve synergistic adaptation by enforcing prediction consistency across modalities.

\begin{figure}[t]
    \centering
    \includegraphics[width=1\linewidth]{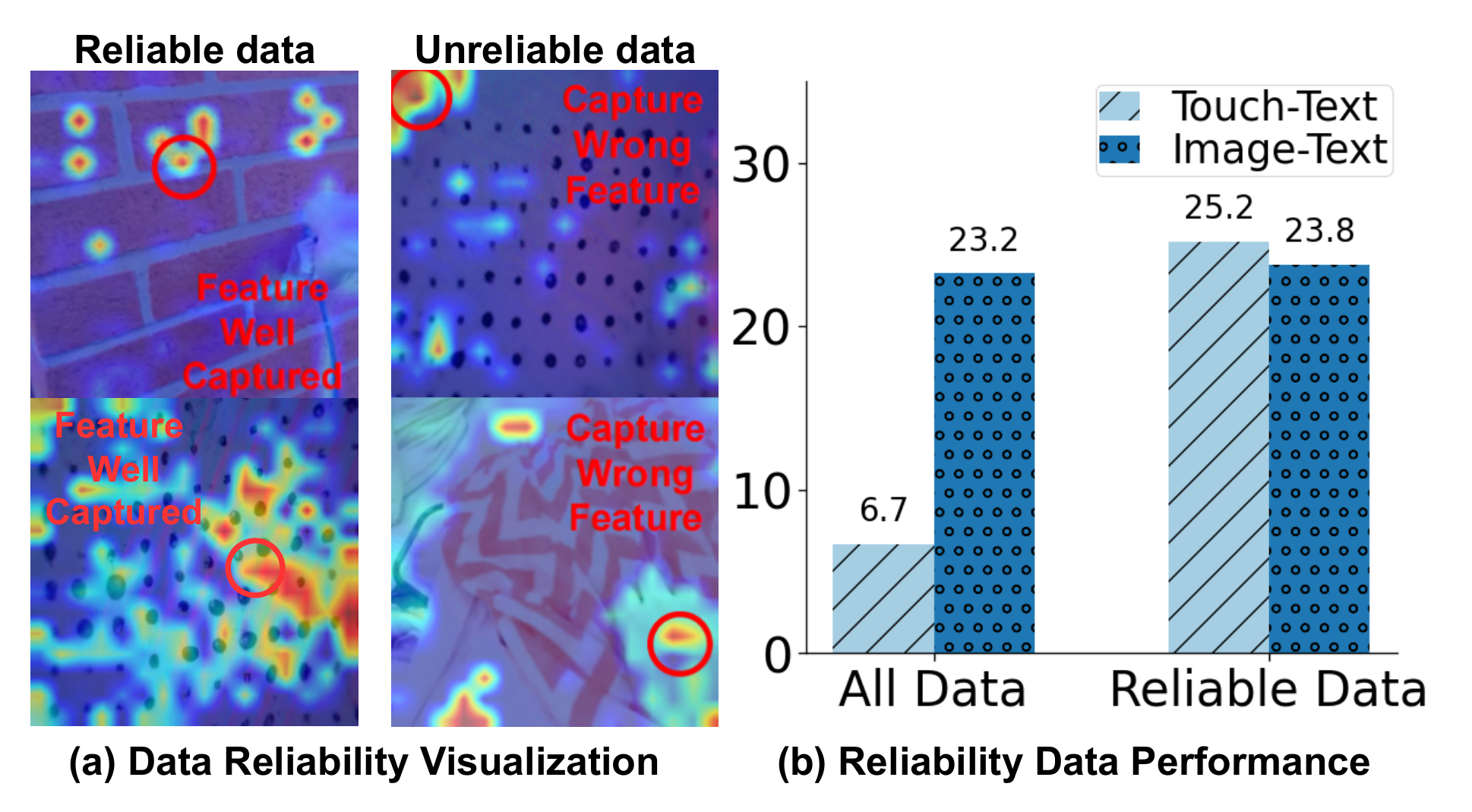}
    \caption{\textbf{Experiments for Data Reliability.} \textbf{(a)} Illustration of High-reliable versus low-reliable samples in multi-modal
    data under distribution shifts by using Grad-CAM. High-reliable samples effectively capture
    relevant features, facilitating robust model adaptation, while low-reliable samples introduce errors by emphasizing irrelevant or missing details. \textbf{(b)} Impact of Data Quality on Material Classification Accuracy Using TENT in TAG-C Datasets. This bar chart compares classification accuracy (\%)
    for Touch-Text and Image-Text modalities when adapting with all available data
    versus only clean data under distribution shifts. Results demonstrate that
    High-reliable (clean) samples significantly boost performance, particularly for
    tactile data, while noisy data in all samples leads to degradation. }
    \label{fig:motivation-Experiments for Data Reliability}
\end{figure}

 However, these existing methods exhibit limitations when confronted with complex, real-world scenarios, severely limiting their effectiveness in practical TVL applications. First, most multimodal TTA methods operate on a fallacious premise: that all test samples are beneficial for adaptation, thus indiscriminately utilizing the entire input stream for updates. This strategy often proves counterproductive when faced with the noisy and corrupted data common in real-world settings. As our analysis reveals (see Figure~\ref{fig:motivation-Experiments for Data Reliability}), \textit{low-reliability} samples where the model focuses on irrelevant background introduce noisy gradients and can cause up to 18.5\% performance drop. Existing methods lack dynamic filtering, resulting in error accumulation from such detrimental samples. 
 Secondly, in real-world TVL environments, ``asynchronous distribution shifts'' often occur, where one modality (e.g., touch) may continually degrade due to physical wear while another (e.g., vision) remains stable. Tactile sensors are especially prone to abrupt signal corruption from factors like wear, surface damage, or contact changes. Notably, simultaneous shifts in all modalities are rare in practice and usually require hardware replacement or environment adjustment. However, existing methods generally use static or simple fusion strategies, such as averaging or fixed weights, which cannot adapt to the real-time reliability of each modality. This limits the model’s robustness when facing asynchronous modality shifts.

To address these gaps, we propose \textbf{RobustTouch}, a reliability-aware framework specifically designed for TVL models under complex, real-world distribution shifts. RobustTouch introduces a dynamic sample filtering mechanism that leverages perturbation-based reliability indicators to automatically and adaptively exclude noisy or corrupted samples from the adaptation process, ensuring that model updates are grounded in genuinely informative data even under severe distribution shifts. Building on this, our quality-aware dynamic fusion module employs the learned robustness signals to assign adaptive, sample-specific weights to each sensory modality, enabling the model to selectively integrate vision and touch cues in a manner that reflects their instantaneous trustworthiness. This selective integration is further reinforced by a reliability-aware loss, which restricts adaptation to high-confidence samples while explicitly promoting both prediction confidence and class diversity, thus preventing error accumulation and model collapse. Our approach makes three key contributions:

\begin{enumerate}
    \item We \textit{first} reveal a new challenge for Tactile-Vision-Language multi-modal test-time adaptation, namely asynchronous distribution shifts, where each sensory modality may experience independent distribution changes. 
    
    \item We propose RobustTouch, a novel TTA algorithm for the TVL model that dynamically evaluates the trustworthiness of each modality to effectively filter out low-quality samples and adaptively balance the contributions across modalities, ensuring robust performance under severe modality degradation.
    

    \item We provide a benchmark, TAG-C, for multi-modal TTA with asynchronous distribution shifts for the TVL applications. Extensive experiments on the benchmarks not only verify the effectiveness of our method but also give some observations for the community.
\end{enumerate}
\section{Related Work}

\textbf{Tactile Multi-modal Perception.}
Integrating tactile sensing with visual and proprioceptive modalities has enabled substantial progress in robotic manipulation and perception \cite{calandra2018more,gadre2021act,li2023see}. Beyond providing complementary cues to vision, tactile feedback contributes to state estimation, contact localization, slip detection, and force inference, which are central to grasp stability and dexterous interaction \cite{bhattacharjee2018multimodal,lin2019learning}. Recent visual-tactile sensors \cite{yuan2017gelsight,donlon2018gelslim,lambeta2020digit} offer high-resolution contact images that support pixel-level reasoning about surface geometry and deformation, making it possible to associate local touch patterns with visual appearance for recognition and manipulation.
With growing sensor diversity and the practical cost of collecting large-scale touch datasets, recent efforts study unified representation learning that bridges tactile, visual, and proprioceptive information \cite{Yang_2024_CVPR,fenganytouch}. Typical pipelines pair tactile images or features with visual embeddings and kinematic signals, and optimize with contrastive or reconstruction-style objectives to align modalities while preserving contact-specific structure. Such representations have been used for object classification, shape reasoning, and contact-rich manipulation, and can facilitate transfer across sensors, objects, and tasks. Together, these directions highlight a trend toward scalable visual-tactile learning, where high-resolution tactile imaging, cross-modal alignment, and unified embeddings provide a foundation for robust perception and control.
\newline
\textbf{Test-Time Adaptation.}
Test-time adaptation (TTA) aims to improve model robustness under distribution shifts by adapting to target data at inference time. Early methods like TTT \cite{sun2020test} and TTT++ \cite{liu2021ttt++} introduce auxiliary self-supervised tasks during training to enable adaptation at test time, but these methods require access to source data and depend heavily on the effectiveness of these tasks. TENT \cite{wangtent} addresses this by optimizing prediction entropy online during testing, thereby eliminating training-time dependency. However, its lack of sample selection makes it vulnerable to unreliable samples, often leading to unstable adaptation. Subsequent works such as CoTTA \cite{wang2022continual}, EATA \cite{niu2022efficient}, SAR \cite{niutowards}, and DeYO \cite{lee2024entropy} further improve the stability of TTA by introducing a sample filtering mechanism that updates model parameters based on reliable samples while augmenting or discarding those with high uncertainty or potential domain mismatch. This design prevents error accumulation during adaptation and enhances the overall robustness of the model under distribution shifts. Nevertheless, these approaches are primarily designed for uni-modal scenarios and do not directly extend to multi-modal settings, where cross-modal interactions are critical. To address this, recent works such as MM-TTA \cite{shin2022mm} employ a mixed pseudo-labeling strategy across 2D and 3D modalities, while READ \cite{yang2024test} reweighs each modality via attention-based reliability estimation. TDA \cite{karmanov2024efficient} enables efficient vision-language model adaptation through training-free dynamic caching and negative pseudo labeling. However, its adaptation capacity is constrained by the limited few-shot cache, which only captures stable distributional patterns and struggles to respond to rapidly changing domains in wild test settings. In contrast, our RobustTouch framework focuses on test-time adaptation for visual–tactile tasks by dynamically filtering out unreliable samples and adaptively fusing modalities according to real-time reliability, rather than relying solely on modality weighting or pseudo labeling.
\begin{figure}[t]
    \centering
    \includegraphics[width=\linewidth]{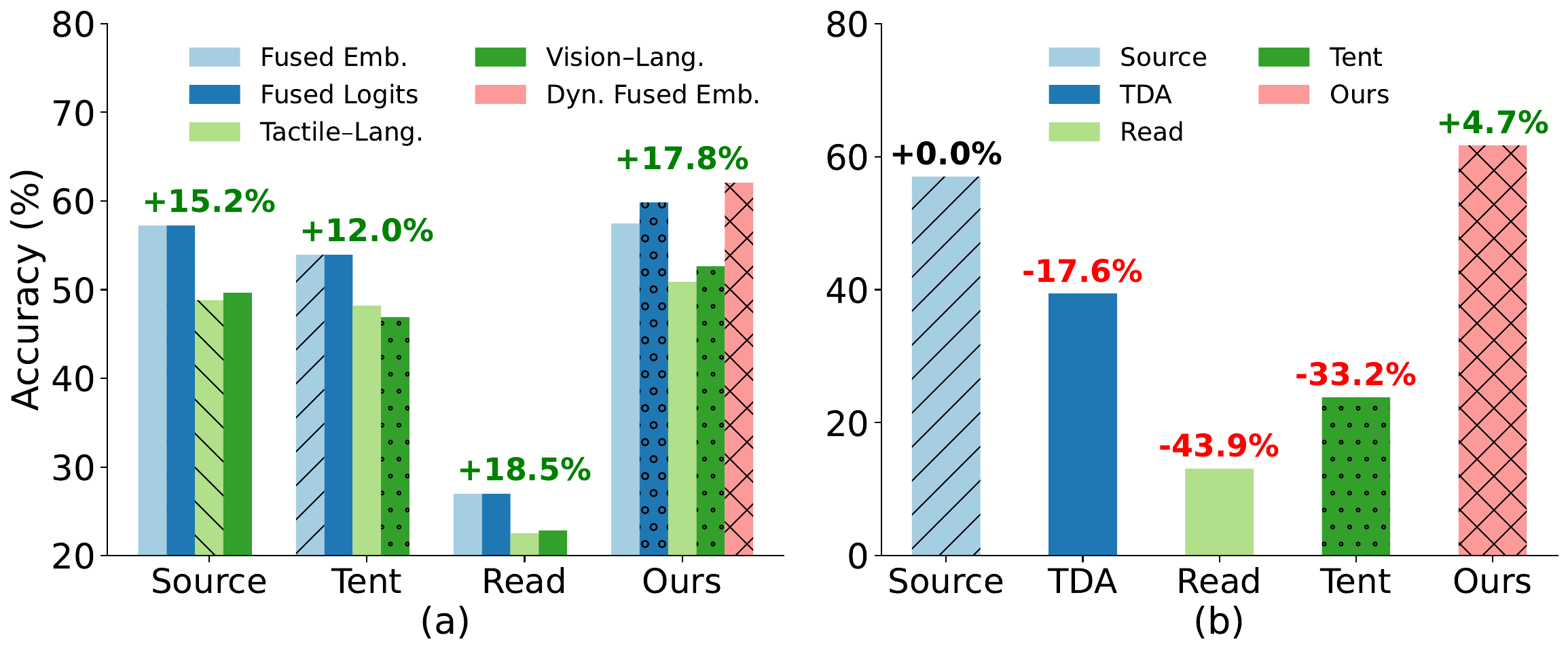}
    \caption{\textbf{Comparison of Accuracy in Material Classification.}  \textbf{(a)} Accuracy across different modalities and fusion methods. Fused Emb. refers to averaging the embeddings of the two modalities, while Fused Logits refers to averaging Tactile-Language and Vision-Language logits. The accuracy gain on the bars is calculated by the difference between the worst performance of single modality accuracy and the best performance in the fusion method.  \textbf{(b)} Accuracy comparison in the material
    classification under continual test-time adaptation scenarios. The source method
    only evaluates the data without adaptation.}
    \label{fig:motivation}
\end{figure}

\section{Formulation and Key Observations}

\subsection{Problem definition}
\textbf{Multi-modal Test-Time Adaptation (MTTA) }extends the concept of single-modality TTA to scenarios involving multiple
sensory modalities, such as tactile, vision, and language.
Formally, let the training data be drawn from the joint distribution $P_{\text{train}}
(\mathcal{X}^{1}, \mathcal{X}^{2}, ..., \mathcal{X}^{M}, y)$, where
$\mathcal{X}^{m}$ denotes the $m$-th modality and $M$ is the total number of
modalities. The model $F_{\Theta_s}$ is trained on this source distribution. At
test time, the data is drawn from a potentially shifted target distribution $P_{\text{test}}
(\mathcal{X}^{1}, \mathcal{X}^{2}, ..., \mathcal{X}^{M})$, where some modalities
may have undergone distribution changes, i.e., $P_{\text{test}}(\mathcal{X}^{m})
\neq P_{\text{train}}(\mathcal{X}^{m})$ for some $m$.

The goal of multi-modal TTA is to adapt the model parameters $\tilde{\Theta}\subseteq
\Theta$ during test time by minimizing an unsupervised objective over the test data:
\begin{equation}
    \min_{\tilde{\Theta}}\mathbf{L}^{\text{tta}}(\mathbf{p}_{m}),
\end{equation}
where $\mathbf{L}^{\text{tta}}$ is typically based on entropy minimization or
consistency regularization, with $m$ indexing different modalities, and
$\mathbf{p}$ represents the model's predictions on the test samples obtained by
the individual modality-specific features, i.e., $\mathbf{p}= C_{\Theta}(f^{1}_{\Theta}
(\mathbf{x}^{1}), ..., f^{M}_{\Theta}(\mathbf{x}^{M}))$.

\textbf{Asynchronous distribution shift in TVL models} refers to the phenomenon where these models, which integrate information from tactile ($\mathcal{T}$), visual ($\mathcal{V}$), and language ($\mathcal{L}$) modalities for robust perception and understanding, encounter modality-specific distribution changes at different times. In real-world robotics and manipulation tasks, it is common that
only one of the sensory modalities (tactile or visual) undergoes a distribution shift
at a given time, due to factors such as lighting changes, sensor wear, or
unexpected physical interactions. We refer to this scenario as \emph{asynchronous
distribution shift}. Formally, let the training data be sampled from the joint distribution $P_{\text{train}}
(\mathcal{T}, \mathcal{V}, \mathcal{L})$. At test time, the model may encounter data
from a shifted distribution, where either the tactile or visual marginal has continuously
changed, but not both simultaneously:
\begin{align}
    P_{\text{test}}(\mathcal{T}', \mathcal{V}, \mathcal{L}) \quad \text{or}\quad P_{\text{test}}(\mathcal{T}, \mathcal{V}', \mathcal{L}),
\end{align}
where $P_{\text{test}}(\mathcal{T}') \neq P_{\text{test}}(\mathcal{T})$ or
$P_{\text{test}}(\mathcal{V}') \neq P_{\text{test}}(\mathcal{V})$, but not both.
This is distinct from the \emph{synchronous shift} setting, where
$P_{\text{test}}(\mathcal{T}', \mathcal{V}', \mathcal{L})$ holds.


\subsection{Key observations}
\subsubsection{High-reliable samples are more beneficial for model adaptation.}



Traditional test-time adaptation updates the model using all test samples, but
this can be detrimental if low-reliable samples are included. Not all inputs are informative—low-reliable
samples may degrade performance rather than improve it. To validate this, we conducted
two experiments: (1) adapting with all samples, and (2) adapting after removing low-reliable samples. As shown in Figure~\ref{fig:motivation-Experiments for Data Reliability}, excluding low-reliable samples yields the best results, while using all
samples leads to performance drops.


Specifically, we identify low-reliable samples based on different criteria for different
modalities. For visual images, low-reliable samples are defined as: (1) defective
or nearly unrecognizable objects due to camera malfunction or damage; and (2) samples
with significant distributional shift from the source domain. For tactile images,
low-reliable samples are defined as: (1) unrecognizable tactile signals caused by
sensor malfunction; and (2) tactile deviations due to insufficient sensor pressure
or inadequate contact duration. Based on these criteria, we selected low-reliable
samples from the original input set.

In addition, we use attention visualization to analyze the regions of focus for
the model when processing these samples. We observe that, for low-reliable
samples, the model tends to focus on background features unrelated to the object,
whereas for High-reliable samples, the model more consistently attends to the object
itself. Therefore, it is necessary to design a method that can precisely assess the
quality of input samples based on the regions attended to by the model, so as to
filter out High-reliable samples for test-time adaptation.

\subsubsection{The inference performance of multi-modal fusion is more robust.}



Existing methods typically perform inference based on a single modality, such as
vision-only or touch-only classification, with little interaction between modalities.
However, we argue that combining both modalities leads to more robust inference.
In real-world settings, input samples are often more complex due to domain shifts
or sensor noise, making unimodal inference unreliable. Since different modalities
can compensate for each other’s shortcomings, multimodal fusion yields superior
performance. To validate this, we conducted three sets of experiments: (1) vision-based
inference, (2) touch-based inference, and (3) multimodal fusion (via result or
logits fusion). As shown in Figure~\ref{fig:motivation}\textbf{(a)}, multimodal fusion
achieves significantly better performance—improving accuracy by over 10\% compared to uni-modal inference.

In the above experiments, we used a fixed fusion ratio. However, in practice,
the optimal ratio should be adaptive. For instance, when visual features become
unreliable, greater emphasis should be placed on tactile features, and vice
versa. Therefore, it is necessary to design a dynamic fusion mechanism that
adjusts the contribution of each modality based on its reliability, further enhancing
the robustness of model inference.

\section{Methodology}
\begin{figure*}
    \centering
    \includegraphics[width=0.9\linewidth]{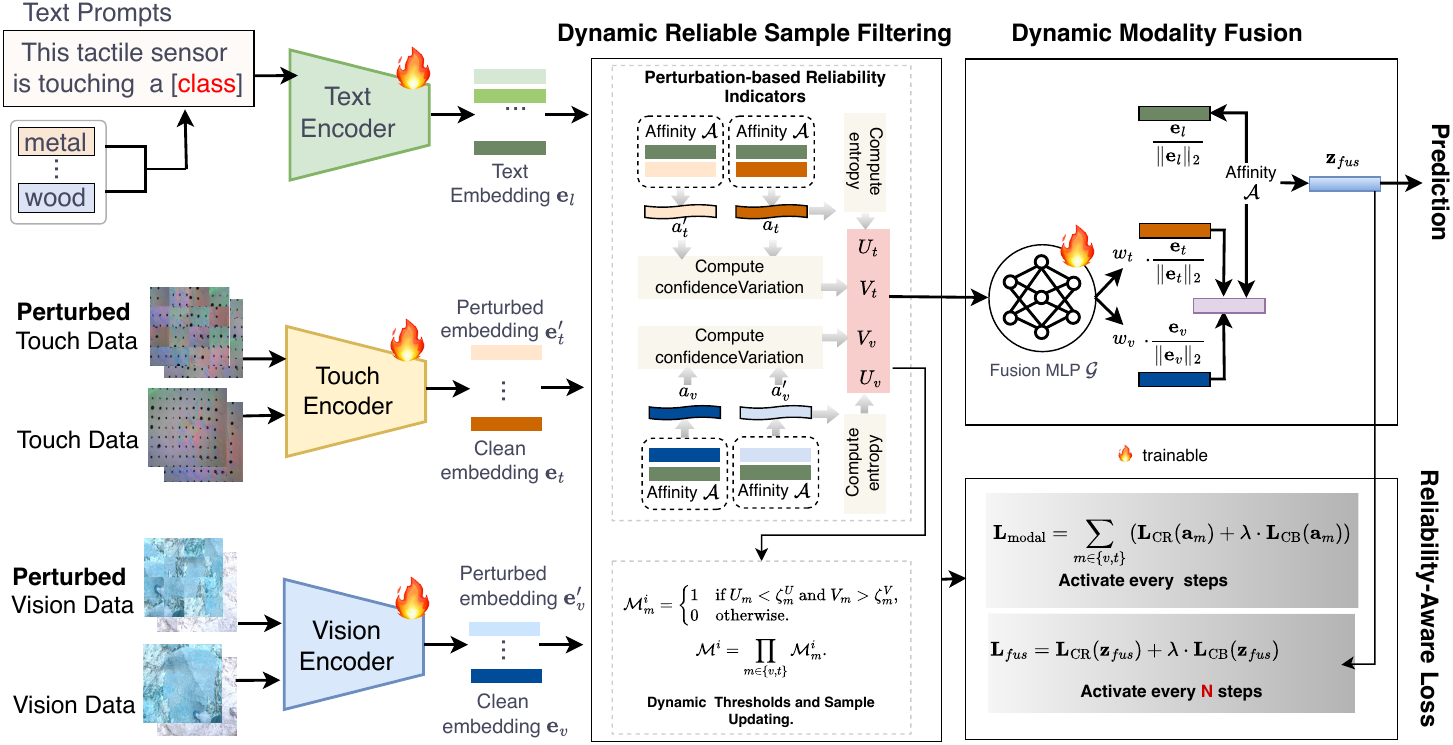}
    \caption{ Overview of the RobustTouch algorithm. The algorithm consists of
    three key modules: (1) Dynamic Reliable Sample Filtering, which identifies
    reliable samples based on perturbation-based reliability indicators; (2) Dynamic
    Modality Fusion, which adaptively combines vision and touch embeddings using
    a lightweight fusion network; (3) Reliability-Aware Loss Design, which
    employs modality-specific losses to drive adaptation. The model updates are focused
    on attention QKV cache and adapters, enabling efficient test-time adaptation
    without requiring labels. }
    \label{fig:placeholder}
\end{figure*}


\subsection{Dynamic Reliable Sample Filtering}

A critical challenge in test-time adaptation is identifying reliable samples amidst
potential distribution shifts. To tackle this, we design a sample filtering module that harnesses perturbation-based reliability metrics. This module produces a binary mask for selecting trustworthy samples during gradient updates and a continuous robustness vector to support downstream adaptive fusion.

\subsubsection{Perturbation-based Reliability Indicators.}
We begin by evaluating how robust each sample is to input perturbation. Specifically,
we define a generic perturbation operator $\text{Perturb}(\cdot)$, which partitions
an input $\mathbf{x}$ into $k$ non-overlapping segments and reassembles them according
to a random permutation.

To quantify the effect of perturbation, we propose an affinity function
$\mathcal{A}(\cdot, \cdot)$ that measures the alignment between raw (unnormalized)
embeddings:
    $\mathcal{A}(\mathbf{e}_{i}, \mathbf{e}_{j}) = \exp(\tau_{\text{aff}}) \cdot \mathbf{e}
    _{i}\mathbf{e}_{j}^{\top} \label{eq:affinity_func}$,
where $\tau_{\text{aff}}$ is a learnable log-temperature parameter,
$\mathbf{e}_{i}$ and $\mathbf{e}_{j}$ are the input embedding (clean or perturbed).
Unlike conventional cosine similarity, our use of unnormalized embeddings
preserves both magnitude and direction, capturing richer information such as feature
strength and model confidence. This enables finer-grained robustness evaluation
and more effective detection of distribution shifts.

For each modality $m \in \{v, t\}$ (vision, touch), we compute the affinity
magnitude for the clean input ($\mathbf{x}_{m}$) and its perturbed version ($\mathbf{x}
'_{m}= \text{Perturb}(\mathbf{x}_{m})$) against the text embeddings
$\mathbf{e}_{l}$. Let $\mathbf{e}_{m}$ and $\mathbf{e}'_{m}$ be the embeddings for
the clean and perturbed input, respectively. $\mathbf{a}_{m}$ and $\mathbf{a}'_{m}$
denote the \textit{affinity vector} between each input (clean or perturbed) and
all text label embeddings. The corresponding affinity vectors are:
    $\mathbf{a}_{m}= \mathcal{A}(\mathbf{e}_{m}, \mathbf{e}_{l}) \quad $and $
    \mathbf{a}'_{m}= \mathcal{A}(\mathbf{e}'_{m}, \mathbf{e}_{l}).$
To minimize overhead, the affinity of the perturbed input ($\mathbf{a}'_{m}$) is
computed with gradients disabled. From this pair of affinity vectors, we derive two
reliability metrics:

\begin{definition}[Prediction Uncertainty]
    Following prior TTA works, we quantify prediction uncertainty by the Shannon entropy of the softmax output over classes. For each modality $m$, given a clean affinity vector $\mathbf{a}_{m}$, its \textbf{Prediction
    Uncertainty} $U_{m}(\mathbf{a}_{m})$ is defined as the entropy of its
    corresponding probability distribution
    $p_{i}= \mathcal{\delta}(\mathbf{a}_{m})$, where $\mathcal{\delta}$ is the softmax
    function, $K$ is the class number:
    \begin{equation}
        U_{m}(\mathbf{a}_{m}) = -\sum_{i=1}^{K}p_{i}\log p_{i}. \label{eq:uncertainty}
    \end{equation}
\end{definition}

\begin{definition}[Confidence Variation]
    For each modality $m$, given the affinity vector from a clean input, $\mathbf{a}
    _{m}$, and its perturbed counterpart, $\mathbf{a}'_{m}$, the \textbf{Confidence
    Variation} $V(\mathbf{a}_{m}, \mathbf{a}'_{m})$ is defined as the difference
    in their prediction confidences:
    \begin{align}
        V_{m}(\mathbf{a}_{m}, \mathbf{a}'_{m}) & = c(\mathbf{a}_{m}) - c(\mathbf{a}'_{m}), \label{eq:variation}
    \end{align}
\end{definition}
The confidence $c(\mathbf{x})$ is
$c(\mathbf{x}) = \max(\mathcal{\delta}(\mathbf{x}))$. For convenience, when referring to the $i$-th sample in a batch, we denote its confidence with the shorthand
$c^{i}\equiv c(\mathbf{a}_{m}^{i})$.

We compute the reliability indicators $(V_{m}, U_{m})$ for each modality $m \in \{
v, t\}$ . These are subsequently concatenated into the final \textit{Robustness
Vector} $\mathcal{r}= [V_{t}, U_{t}, V_{v}, U_{v}]$.


\subsubsection{Dynamic Thresholds and Sample Updating.}
A key innovation of our approach is the use of adaptive, batch-wise thresholds, which
are dynamically updated to reflect the statistical characteristics of each mini-batch.
This allows our filtering to adjust to varying levels of batch difficulty and
label noise.

To determine the dynamic thresholds for each modality $m$, we leverage the 3-sigma
rule from statistics, which adapts to the distribution properties of each batch
and defines $\zeta^{U}_{m}$ as the uncertainty threshold and $\zeta^{V}_{m}$ as
the confidence variation threshold:
\begin{equation}
    \zeta^{U}_{m}= \mu^{U}_{m}+ \alpha \cdot \sigma^{U}_{m}, \quad \zeta_{m}^{V}=
    \mu^{V}_{m}- \alpha \cdot \sigma^{V}_{m},
\end{equation}
where $\mu^{U}_{m}$ and $\sigma^{U}_{m}$ are the mean and standard deviation of uncertainty
values across the batch, while $\mu^{V}_{m}$ and $\sigma^{V}_{m}$ are the mean
and the standard deviation of confidence variation values. $\alpha$ is the hyperparameter
to control the filtration strength. This approach allows thresholds to
automatically adjust based on batch characteristics---becoming stricter when the batch shows consistent behavior and is more lenient when there's greater variation.

A modality is considered reliable if its uncertainty and confidence variation are simultaneously below their respective thresholds. Hence, for each sample $i$ in each modality $m$, we define a binary reliability mask:
\begin{equation}
    \mathcal{M}_{m}^{i}=
    \begin{cases}
        1 & \text{if }U_{m}< \zeta^{U}_{m}\text{ and }V_{m}> \zeta^{V}_{m}, \\
        0 & \text{otherwise}.
    \end{cases}
    \label{eq:modality_mask}
\end{equation}

A sample is considered globally reliable only if all modalities are reliable.
The final mask for sample $i$, defined as $\mathcal{M}^{i}$, is therefore
the logical AND of the individual modality masks, which can be expressed
compactly as their product:
\begin{equation}
    \mathcal{M}^{i}= \prod_{m \in \{v,t\}}\mathcal{M}_{m}^{i}. \label{eq:final_mask_product}
\end{equation}

During model update, unreliable samples (as indicated by $\mathcal{M}^{i}$) are excluded
by detaching their logits from the computation graph. This conservative update
strategy ensures that model adaptation is driven exclusively by high-confidence,
cross-modal-consistent signals, minimizing the risk of propagating noisy or
erroneous pseudo-labels.

\subsection{Dynamic Modality Fusion}
Our dynamic fusion module leverages embeddings from multiple modalities (vision and
touch), which are fused by a lightweight network $\mathcal{G}(\cdot)$. This
network predicts modality-specific weights for each sample based on its
robustness indicator $\mathcal{r}$, allowing the model to dynamically prioritize
the more reliable modality.


Specifically, a small multi-layer perceptron (MLP) $\mathcal{G}$, maps the
robustness indicator $\mathcal{r}$ to a pair of fusion weights $[w_{v}, w_{t}]$
for vision and touch modality:
\begin{equation}
    \mathbf{w}= [w_{v}, w_{t}] = \mathcal{G}(\mathcal{r}), w_{v}, w_{t}\in \mathbb{R}
    .
\end{equation}

The final fused embedding, $\mathbf{e}_{fus}$, is then computed as a convex
combination of the $\ell_{2}$-normalized embeddings:
\begin{equation}
    \mathbf{e}_{fus}= w_{v}\frac{\mathbf{e}_{v}}{\|\mathbf{e}_{v}\|_{2}}+ w_{t}\frac{\mathbf{e}_{t}}{\|\mathbf{e}_{t}\|_{2}}
    . \label{eq:fusion}
\end{equation}

Finally, for zero-shot classification, we compute the fused logits
$\mathbf{z}_{fus}\in \mathbb{R}^{C}$ by applying the affinity function
$\mathcal{A}(\cdot, \cdot)$ between the fused embedding $\mathbf{e}_{fus}$ and
the set of text embeddings $\mathbf{e}_{l}$:
\begin{equation}
    \mathbf{z}_{fus}= \mathcal{A}(\mathbf{e}_{fus}, \frac{\mathbf{e}_{l}}{\|\mathbf{e}_{l}\|_{2}}
    ). \label{eq:logit_fus}
\end{equation}

\subsection{Reliability-Aware Loss}
To drive effective adaptation, we introduce a reliability-aware loss framework that operates in two distinct stages. This entire process is governed by a dual-optimization strategy, ensuring that both the modality-specific branches and the fusion module adapt in a stable and balanced manner. Details as follows:



To guide the adaptation process, our loss functions are computed exclusively on reliable
samples. We formally define the set of reliable samples $\mathcal{S}$ for the
current mini-batch as those indices $i$ for which the per-sample reliability
mask $\mathcal{M}^{i}$ is active, defined as $\mathcal{S}= \{i \mid \mathcal{M}^{i}
= 1 \}$. All subsequent loss functions are computed over this set S, ensuring that
model updates are driven only by high-quality signals.

The first component is a confidence-regularized loss $ \mathbf{L}_{\text{CR}}$ that
encourages high-confidence predictions while penalizing overconfidence. Formally,
given a mini-batch of predictions, the loss function for modality $m$ is
designed as :
\begin{equation}
     \mathbf{L}_{\text{CR}}(\mathbf{a}_{m})= \frac{1}{|\mathcal{S}|}\sum_{i \in
    \mathcal{S}}c^{i}\log \left( \frac{e^{\gamma}}{c^{i}}\right ),
\end{equation}
where the \textbf{$\gamma$} is confidence-regulating hyperparameter.

To encourage prediction diversity, we introduce the Class-Balanced Loss for
modality $m$, $ \mathbf{L}_{\text{CB}}$, which operates on a subset of reliable
samples (indexed by $\mathcal{S}$) within each batch. It is defined as the negative
entropy of a batch-level class distribution, $\hat{\mathbf{p}}$, which is formed
by summing the probabilities of reliable samples and re-normalizing the result:
\begin{equation}
     \mathbf{L}_{\text{CB}}(\mathbf{a}_{m}) = \hat{\mathbf{p}}\cdot \log \hat{\mathbf{p}}
    ,\quad \text{where}\quad \hat{\mathbf{p}}= \mathcal{\delta}\left(\sum_{i \in
    \mathcal{S}}\mathcal{\delta}(\mathbf{a}_{m}^{i})\right). \label{eq:lcb}
\end{equation}
Here, $\mathbf{a}_{m}^{i}$ are the output logits for the $i$-th sample. The
vector $\hat{\mathbf{p}}\in \mathbb{R}^{K}$ represents the normalized class
distribution aggregated from all reliable samples in the batch, with $K$ being the
total number of classes. The log operation is applied element-wise.

Our adaptation framework employs a distinct, two-pronged optimization strategy for
the modality-specific branches and the fusion module.

The modality-specific branches ($m \in \{v, t\}$) are guided at each training step.
This is achieved through the modality adaptation loss, $\mathbf{L}_{\text{modal}}$,
which is computed on the logits $\mathbf{a}_{m}$ from each branch:
\begin{equation}
     \mathbf{L}_{\text{modal}}= \sum_{m \in \{v, t\}}\left(  \mathbf{L}_{\text{CR}}
    (\mathbf{a}_{m}) + \lambda \cdot  \mathbf{L}_{\text{CB}}(\mathbf{a}_{m})\right
    ),
\end{equation}
where $\lambda$ is a hyperparameter to control the strength of the class-balanced
loss.

The dynamic fusion loss, $ \mathbf{L}_{\text{fus}}$, is handled by a separate optimizer
with a delayed update schedule. To obtain a more stable training signal, gradients
are accumulated over $N$ batches (e.g., $N=5$), and the optimizer step is
performed only once. The resulting aggregated gradient provides a more stable and
comprehensive training signal, which is crucial for robustly tuning the
sensitive parameters of the fusion mechanism. The loss is defined as:
\begin{equation}
    \mathbf{L}_{fus}= \mathbf{L}_{\text{CR}}(\mathbf{z}_{fus}) + \lambda \cdot
    \mathbf{L}_{\text{CB}}(\mathbf{z}_{fus}).
\end{equation}

Crucially, both loss computations are performed exclusively on reliable samples,
ensuring that the adaptation for both the individual modalities and their fusion
is driven by high-quality signals.

\section{Experiments}
\begin{table*}[ht!]
\small
\centering
\caption{\textbf{Top-1} accuracy (\%) comparison of \textbf{RobustTouch} with TTA baselines on \textbf{TAG‑C} benchmark (corrupted visual modality) under the continuous cross-domain setting.}
\begin{tabular}{l|lllllllllllllll|c}\hline
Time & \multicolumn{15}{l|}{$t\xrightarrow{\hspace*{14cm}}$}& \\ \hline
Methods
  & \rotatebox[origin=c]{70}{Brit.}
  & \rotatebox[origin=c]{70}{Defoc.}
  & \rotatebox[origin=c]{70}{Fog}
  & \rotatebox[origin=c]{70}{Gauss.}
  & \rotatebox[origin=c]{70}{Impul.}
  & \rotatebox[origin=c]{70}{Motion}
  & \rotatebox[origin=c]{70}{Shot}
  & \rotatebox[origin=c]{70}{Contr.}
  & \rotatebox[origin=c]{70}{Elastic}
  & \rotatebox[origin=c]{70}{Frost}
  & \rotatebox[origin=c]{70}{Glass}
  & \rotatebox[origin=c]{70}{JPEG}
  & \rotatebox[origin=c]{70}{Pix.}
  & \rotatebox[origin=c]{70}{Snow}
  & \rotatebox[origin=c]{70}{Zoom}
  & Avg \\
\hline

No Adapt.& 61.0&	61.3&	50.2&	58.9&	60.6&	59.4&	60.9&	57.2&	62.6&	48.8&	57.2&	62.2&	62.8	&49.2&	55.7&	57.9\\
TENT & 31.7    & 23.1 & 23.1   & 23.1    & 23.1  & 23.1   & 23.1 & 23.1 & 23.1  & 23.1 & 23.1       & 23.1     & 23.1           & 23.1     & 23.1 & 23.7 \\
SAR &61.0&	61.3&	50.2&	58.9&	60.6&	59.4&	60.8&	57.1&	62.6&	48.9&	57.3&	62.2&	62.8&	49.2&	55.5&	57.8\\
TDA  & 43.1     & 41.8 & 21.2    & 30.2    & 34.9  & 40.2   & 36.7 & 27.4 & 35.0   & 15.5 & 33.5       & 36.1     &38.9          & 17.8     & 29.2 & 32.1 \\
READ & 34.5 & 13.4 & 11.3    & 10.6    & 10.2  & 10.1  & 9.8 & 10.0 & 9.8   & 10.0 & 9.5      & 9.7     & 9.7           & 9.3     & 9.3 & 11.8 \\\hline
\textbf{RobustTouch}      & \textbf{61.6}&	\textbf{65.0}&	\textbf{51.2}&	\textbf{62.3}&	\textbf{64.8}&	\textbf{66.7}&	\textbf{68.6}&	\textbf{67.7}&	\textbf{70.5}&	\textbf{55.6}&	\textbf{57.7}&	\textbf{60.5}&	\textbf{62.4}&	\textbf{53.0}&	\textbf{57.7}&	\textbf{61.7}\\\hline
\end{tabular}
\vspace{-1mm}

\label{tab:ctta_tag}
\end{table*}

\subsection{Experiment Settings}
\textbf{Benchmarks.}
We construct a benchmark based on Touch and GO (TAG) \cite{yang2022touch}—denoted TAG‑C. TAG-C comprises two independent corruption paradigms designed to reflect real-world failure modes:

\textit{Visual Corruptions (15 types):} We adopt the established corruption suite from ImageNet-C \cite{hendrycks2019benchmarking}, covering four categories: Noise, blur, Weather, and Digital. This choice ensures our visual modality evaluation aligns with established TTA methods (e.g., TENT \cite{wangtent}, SAR \cite{niutowards} and READ \cite{yang2024test}).

\textit{Tactile Corruptions (7 types):} Vision-based tactile sensors (e.g., Gelsight \cite{yuan2017gelsight}) use enclosed overhead cameras to image elastomer deformation. To capture realistic degradation patterns, we adapt ImageNet-C by retaining only corruptions that correspond to actual failure mechanisms in these sensors and exclude corruptions such as weather-related corruptions (e.g., Fog, Frost, Snow).

Each type of corruption has 5 severity levels. In our experiments, we evaluate at severity level 3. Following the established practice of ImageNet-C and recent multi-modal TTA benchmarks, we employ synthetic corruptions to enable reproducible, controlled evaluation. We aim to emulate the distribution shifts that Tactile‑Vision‑Language models are likely to encounter in real‑world scenarios. Further benchmark details can be found in the Appendix.
\newline
\textbf{Evaluation Scenarios.}
Based on the TAG-C benchmark, we designed two evaluation scenarios for \textbf{Asynchronous distribution shift} in TVL models to emulate real-world conditions. (1) \textit{Continuous cross-domain setting}, presents the model with homogeneous batches, where each batch belongs to a single corruption domain and the domains shift sequentially (e.g., A→B→C). This tests adaptation to systematic changes. (2)\textit{ The dynamic wild setting}, involves continuous random mixed-domain transitions. Here, each batch comprises a heterogeneous mix of samples randomly drawn from multiple corruption domains, better reflecting the stochastic nature of real-world data streams. Further experimental details are available in the Appendix.
\newline
\textbf{Baselines.}
We compare RobustTouch against several state-of-the-art TTA approaches, including both single-modal and multi-modal methods, to comprehensively evaluate our contribution:

\textit{Single-Modal TTA Methods.} These methods were originally designed for single-modality adaptation (typically vision) and do not explicitly model multi-modal interactions: TENT \cite{wangtent}, SAR \cite{niutowards}.

\textit{Multi-Modal TTA Methods.} These methods explicitly address distribution shifts in multi-modal settings by modeling cross-modal interactions and coordinating adaptation across modalities, making them our primary competitors: TDA \cite{karmanov2024efficient}, and READ \cite{yang2024test}.
\newline
\textbf{Implementation Details.} The experiments are run on the Ubuntu 22.04 platform with Nvidia RTX 4090 GPUs. As our work focuses on test-time adaptation rather than pretraining, we employ the publicly available pretrained CLIP model \cite{cherti2023reproducible} from Anytouch \cite{fenganytouch}. This allows us to isolate the contributions of our TTA framework from pretraining design choices. During the test time, RobustTouch uses the AdamW optimizer \cite{loshchilov2017decoupled} for optimization, with the learning rate of $1\times10^{-6}$ within a single epoch. The $\alpha$ in the dynamic thresholds and sample updating is 1 for all experiments. The $\lambda$ is 0.5 in the reliability-aware loss function for all experiments. Also, confidence-regulating hyperparameter $\gamma$ is $e^{-1}$.

\subsection{Main Results}
We evaluate the performance of our proposed RobustTouch method by comparing it against baselines under a continuous domain variation scenario. Table \ref{tab:ctta_tag}–\ref{tab:ctta_gel}  respectively presents the experimental results of RobustTouch and the compared baselines under visual and tactile corruptions on TAG-C. From the empirical findings, the following key insights and observations emerge.

\begin{table*}[t!]
  \centering
  \caption{\textbf{Top-1} accuracy (\%) comparison of \textbf{RobustTouch} with TTA baselines on \textbf{TAG-C} benchmark (corrupted visual modality) under the dynamic wild setting (continuous random mixed-domain).}
  \begin{tabular}{l|cccccc}
    \toprule
    Methods & No Adapt. & TENT & SAR & TDA & READ & \textbf{RobustTouch} \\
    \midrule
    Accuracy (\%) & 57.9 & 23.7 & 57.8 & 32.8 & 13.1 & \textbf{62.0} \\
    \bottomrule
  \end{tabular}

  \label{tab:dynamic_video}
\end{table*}

\begin{table*}[t!]
\centering
\caption{\textbf{Top-1} accuracy (\%) comparison of \textbf{RobustTouch} with TTA baselines on \textbf{TAG-C} benchmark (corrupted tactile modality) under the continuous cross-domain setting.}
\begin{tabular}{l|ccccccc|c}
\hline
Time 
  & \multicolumn{7}{l|}{$t\xrightarrow{\hspace*{8.5cm}}$}  \\
\hline
Methods
  & \rotatebox[origin=c]{0}{Brit.}
  & \rotatebox[origin=c]{0}{Defoc.}
  & \rotatebox[origin=c]{0}{Gauss.}
  & \rotatebox[origin=c]{0}{Impul.}
  & \rotatebox[origin=c]{0}{Motion}
  & \rotatebox[origin=c]{0}{Contr.}
  & \rotatebox[origin=c]{0}{Elastic}
  & Avg
\\
\hline
No Adapt.   
  & 63.0 & 62.7 & 55.4 & 60.0 & 58.8 & 62.2& 62.7&60.7
   \\
TENT        
  & 34.4 & 23.1 & 23.1 & 23.1 & 23.1 & 23.1& 23.1&24.7
   \\
SAR         
  & 63.0 & 62.7 & 55.2 & 59.9 & 58.7 & 62.2& 62.7&60.6
   \\
TDA         
  & 39.2 & 29.5 & 29.5 & 28.2 & 28.2 & 28.2& 28.2&30.1
   \\
READ        
  & 34.7 & 14.3 & 13.8 & 14.1 & 10.8 & 13.0& 11.7&16.1
  \\\hline
\textbf{RobustTouch} 
  & \textbf{67.6} & \textbf{69.0} & \textbf{64.2} & \textbf{71.5} & \textbf{65.7} & \textbf{66.2} & \textbf{69.0} & \textbf{67.6}
  \\
\hline
\end{tabular}
\vspace{-1mm}

\label{tab:ctta_gel}
\end{table*}

\noindent\textbf{Comparison with State-of-the-art.}
The experimental evaluation on the TAG-C benchmark demonstrates the superior performance of RobustTouch across multiple corruption scenarios and modality configurations. Under the continuous cross-domain setting with corrupted video modality (Table~\ref{tab:ctta_tag}), RobustTouch achieves an average accuracy of 61.7\%, substantially outperforming all baseline methods and surpassing the strongest baseline by 3.8\%.

The advantage of RobustTouch is further amplified in the continuous cross-domain setting with corrupted tactile modality (Table~\ref{tab:ctta_gel}), where it achieves its largest margin, reaching 67.6\% average accuracy---a substantial 6.9\% improvement over the strongest baseline (No Adapt.). Notably, RobustTouch sustains this performance, whereas other adaptation methods, such as TDA and READ experience dramatic drops to 30.1\% and 16.1\%, respectively.

The dynamic wild setting (Table \ref{tab:dynamic_video}) severely tests the limits of TTA by introducing continuous, random domain shifts. In this demanding scenario, RobustTouch demonstrates remarkable resilience, achieving 62.0\% accuracy. This performance stands in stark contrast to all baselines, surpassing the runner-up by 4.1 percentage points and avoiding the catastrophic performance degradation seen in methods like READ (13.1\%).

\begin{figure}
    \centering
    \includegraphics[width=1\linewidth]{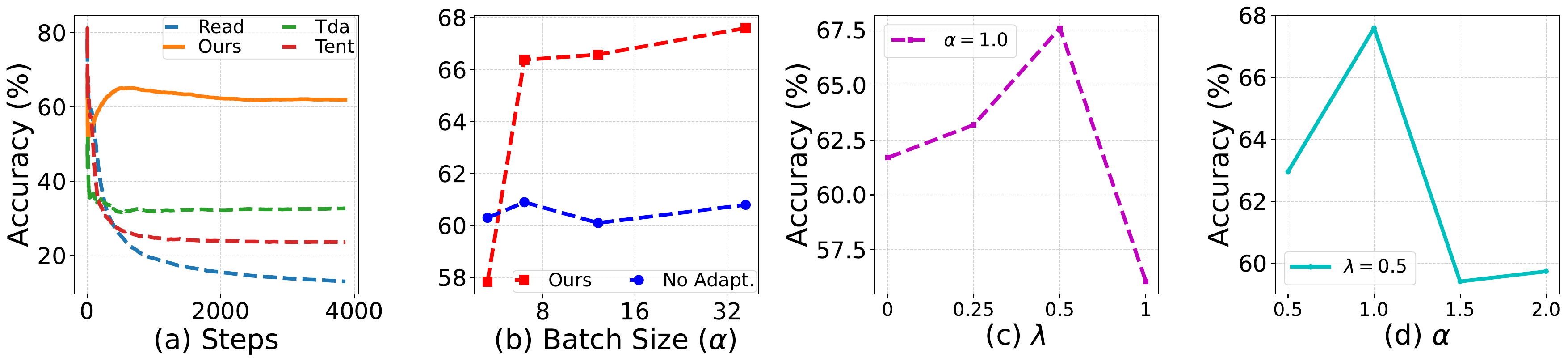}
    \caption{(a) Performance Trend over Time under the dynamic wild setting on TAG-C benchmark (corrupted visual modality). (b) Sensitivity test of batch size in tactile modality corruption on TAG-C benchmark (corrupted tactile modality). (c) Sensitivity test of different $\lambda$ on TAG-C benchmark (corrupted tactile modality). (d) Sensitivity test of different $\alpha$ on TAG-C benchmark (corrupted tactile modality).}
    \label{fig:Analytic Studies}
\end{figure}

\subsection{Ablation and Analytic Studies}
\textbf{Ablation studies.} Our ablation study on the TAG-C benchmark under a continuous cross-domain setting, considering visual and tactile corruptions separately, as shown in Table~\ref{tab:ablation-k}\begin{table}[t]
\setlength{\tabcolsep}{3pt}
\small
\centering
\caption{Ablation study (\textbf{Top-1} accuracy (\%)).}
\begin{tabular}{c c c |c c}
\toprule
\makecell{Dynamic\\Filter} & \makecell{Dynamic\\Fusion} & \makecell{Reliability-\\Aware Loss} & \makecell{Visual\\Corruption} & \makecell{Tactile\\Corruption} \\
\midrule
\ding{51} & \ding{51} & \ding{55} & 57.6 & 64.5 \\
\ding{51} & \ding{55} & \ding{51} & 59.0 & 59.4 \\
\ding{55} & \ding{51} & \ding{51} & 59.3 & 63.2 \\
\ding{51} & \ding{51} & \ding{51} & \textbf{61.7} & \textbf{67.6} \\

\bottomrule
\end{tabular}

\label{tab:ablation-k}
\end{table}
, validates the synergy of our framework's components against the full model's performance of 61.7\% (visual) and 67.6\% (tactile) accuracy. Each module plays a critical, non-redundant role. The absence of the Reliability-Aware Loss is particularly damaging under visual corruption, reducing accuracy to 57.6\%. Meanwhile, removing Dynamic Fusion most severely impacts the tactile stream, causing accuracy to fall to 59.4\%. This underscores that the concerted action of all three components is necessary for SOTA robustness (Details in the Appendix).
\newline
\textbf{Analytic studies.}
We further analyze the characteristics of our method. First, as shown in Table \ref{fig:Analytic Studies}(a), it demonstrates the model's capacity for effective online adaptation. Second, as shown in Table \ref{fig:Analytic Studies}(b), we investigate the model's sensitivity to batch size and find that its performance remains stable across various configurations (Details and other sensitivity tests of hyperparameters in the Appendix).

\section{Conclusion}
We present RobustTouch, a novel test-time adaptation framework designed to overcome asynchronous distribution shifts through dynamic sample filtering and adaptive multi-modal fusion. Extensive evaluation of our proposed TAG-C benchmark demonstrates a significant performance gain over SOTA methods in multiple complex settings, underscoring the criticality of reliability-aware adaptation. We hope RobustTouch and TAG-C offer useful foundations for future work.

{
    \small
    \bibliographystyle{ieeenat_fullname}
    \bibliography{main}

@String(CVPR= {IEEE Conf. Comput. Vis. Pattern Recog.})

@String(AAAI = {AAAI})

@String(CVPR  = {CVPR})

@article{calandra2018more,
  title={More than a feeling: Learning to grasp and regrasp using vision and touch},
  author={Calandra, Roberto and Owens, Andrew and Jayaraman, Dinesh and Lin, Justin and Yuan, Wenzhen and Malik, Jitendra and Adelson, Edward H and Levine, Sergey},
  journal={IEEE Robotics and Automation Letters},
  volume={3},
  number={4},
  pages={3300--3307},
  year={2018},
  publisher={IEEE}
}

@inproceedings{gadre2021act,
  title={Act the part: Learning interaction strategies for articulated object part discovery},
  author={Gadre, Samir Yitzhak and Ehsani, Kiana and Song, Shuran},
  booktitle={Proceedings of the IEEE/CVF International Conference on Computer Vision},
  pages={15752--15761},
  year={2021}
}

@inproceedings{li2023see,
  title={See, Hear, and Feel: Smart Sensory Fusion for Robotic Manipulation},
  author={Li, Hao and Zhang, Yizhi and Zhu, Junzhe and Wang, Shaoxiong and Lee, Michelle A and Xu, Huazhe and Adelson, Edward and Fei-Fei, Li and Gao, Ruohan and Wu, Jiajun},
  booktitle={Conference on Robot Learning},
  pages={1368--1378},
  year={2023},
  organization={PMLR}
}

@article{yuan2017gelsight,
  title={Gelsight: High-resolution robot tactile sensors for estimating geometry and force},
  author={Yuan, Wenzhen and Dong, Siyuan and Adelson, Edward H},
  journal={Sensors},
  volume={17},
  number={12},
  pages={2762},
  year={2017},
  publisher={MDPI}
}

@inproceedings{donlon2018gelslim,
  title={Gelslim: A high-resolution, compact, robust, and calibrated tactile-sensing finger},
  author={Donlon, Elliott and Dong, Siyuan and Liu, Melody and Li, Jianhua and Adelson, Edward and Rodriguez, Alberto},
  booktitle={2018 IEEE/RSJ International Conference on Intelligent Robots and Systems (IROS)},
  pages={1927--1934},
  year={2018},
  organization={IEEE}
}

@article{lambeta2020digit,
  title={Digit: A novel design for a low-cost compact high-resolution tactile sensor with application to in-hand manipulation},
  author={Lambeta, Mike and Chou, Po-Wei and Tian, Stephen and Yang, Brian and Maloon, Benjamin and Most, Victoria Rose and Stroud, Dave and Santos, Raymond and Byagowi, Ahmad and Kammerer, Gregg and others},
  journal={IEEE Robotics and Automation Letters},
  volume={5},
  number={3},
  pages={3838--3845},
  year={2020},
  publisher={IEEE}
}

@article{bhattacharjee2018multimodal,
  title={Multimodal tactile perception of objects in a real home},
  author={Bhattacharjee, Tapomayukh and Clever, Henry M and Wade, Joshua and Kemp, Charles C},
  journal={IEEE Robotics and Automation Letters},
  volume={3},
  number={3},
  pages={2523--2530},
  year={2018},
  publisher={IEEE}
}

@inproceedings{lin2019learning,
  title={Learning to identify object instances by touch: Tactile recognition via multimodal matching},
  author={Lin, Justin and Calandra, Roberto and Levine, Sergey},
  booktitle={2019 International Conference on Robotics and Automation (ICRA)},
  pages={3644--3650},
  year={2019},
  organization={IEEE}
}

@InProceedings{Yang_2024_CVPR,
    author    = {Yang, Fengyu and Feng, Chao and Chen, Ziyang and Park, Hyoungseob and Wang, Daniel and Dou, Yiming and Zeng, Ziyao and Chen, Xien and Gangopadhyay, Rit and Owens, Andrew and Wong, Alex},
    title     = {Binding Touch to Everything: Learning Unified Multimodal Tactile Representations},
    booktitle = {Proceedings of the IEEE/CVF Conference on Computer Vision and Pattern Recognition (CVPR)},
    month     = {June},
    year      = {2024},
    pages     = {26340-26353}
}

@inproceedings{fenganytouch,
  title={AnyTouch: Learning Unified Static-Dynamic Representation across Multiple Visuo-tactile Sensors},
  author={Feng, Ruoxuan and Hu, Jiangyu and Xia, Wenke and Shen, Ao and Sun, Yuhao and Fang, Bin and Hu, Di and others},
  booktitle={The Thirteenth International Conference on Learning Representations},
  year={2025}
}

@inproceedings{sun2020test,
  title={Test-time training with self-supervision for generalization under distribution shifts},
  author={Sun, Yu and Wang, Xiaolong and Liu, Zhuang and Miller, John and Efros, Alexei and Hardt, Moritz},
  booktitle={International conference on machine learning},
  pages={9229--9248},
  year={2020},
  organization={PMLR}
}

@article{liu2021ttt++,
  title={Ttt++: When does self-supervised test-time training fail or thrive?},
  author={Liu, Yuejiang and Kothari, Parth and Van Delft, Bastien and Bellot-Gurlet, Baptiste and Mordan, Taylor and Alahi, Alexandre},
  journal={Advances in Neural Information Processing Systems},
  volume={34},
  pages={21808--21820},
  year={2021}
}

@inproceedings{guosmoothing,
  title={Smoothing the Shift: Towards Stable Test-Time Adaptation under Complex Multimodal Noises},
  author={Guo, Zirun and Jin, Tao},
  year         = 2025,
  booktitle={The Thirteenth International Conference on Learning Representations}
}

@inproceedings{wangtent,
  title={Tent: Fully Test-Time Adaptation by Entropy Minimization},
  author={Wang, Dequan and Shelhamer, Evan and Liu, Shaoteng and Olshausen, Bruno and Darrell, Trevor},
  year         = 2021,
  booktitle={International Conference on Learning Representations}
}

@inproceedings{wang2022continual,
  title={Continual test-time domain adaptation},
  author={Wang, Qin and Fink, Olga and Van Gool, Luc and Dai, Dengxin},
  booktitle={Proceedings of the IEEE/CVF Conference on Computer Vision and Pattern Recognition},
  pages={7201--7211},
  year={2022}
}

@inproceedings{niu2022efficient,
  title={Efficient test-time model adaptation without forgetting},
  author={Niu, Shuaicheng and Wu, Jiaxiang and Zhang, Yifan and Chen, Yaofo and Zheng, Shijian and Zhao, Peilin and Tan, Mingkui},
  booktitle={International conference on machine learning},
  pages={16888--16905},
  year={2022},
  organization={PMLR}
}

@inproceedings{niutowards,
  title={Towards Stable Test-time Adaptation in Dynamic Wild World},
  author={Niu, Shuaicheng and Wu, Jiaxiang and Zhang, Yifan and Wen, Zhiquan and Chen, Yaofo and Zhao, Peilin and Tan, Mingkui},
  year={2023},
  booktitle={ International Conference on Learning Representations}
}

@inproceedings{lee2024entropy,
  title={Entropy is not enough for test-time adaptation: From the perspective of disentangled factors},
  author={Lee, Jonghyun and Jung, Dahuin and Lee, Saehyung and Park, Junsung and Shin, Juhyeon and Hwang, Uiwon and Yoon, Sungroh},
  booktitle={ International Conference on Learning Representations},
  year={2024}
}

@inproceedings{shin2022mm,
  title={Mm-tta: multi-modal test-time adaptation for 3d semantic segmentation},
  author={Shin, Inkyu and Tsai, Yi-Hsuan and Zhuang, Bingbing and Schulter, Samuel and Liu, Buyu and Garg, Sparsh and Kweon, In So and Yoon, Kuk-Jin},
  booktitle={Proceedings of the IEEE/CVF Conference on Computer Vision and Pattern Recognition},
  pages={16928--16937},
  year={2022}
}

@inproceedings{yang2024test,
  title={Test-time adaptation against multi-modal reliability bias},
  author={Yang, Mouxing and Li, Yunfan and Zhang, Changqing and Hu, Peng and Peng, Xi},
  booktitle={International Conference on Learning Representations},
  year={2024}
}

@article{yang2022touch,
  title={Touch and Go: Learning from Human-Collected Vision and Touch},
  author={Yang, Fengyu and Ma, Chenyang and Zhang, Jiacheng and Zhu, Jing and Yuan, Wenzhen and Owens, Andrew},
  journal={Advances in Neural Information Processing Systems},
  volume={35},
  pages={8081--8103},
  year={2022}
}

@article{karmanov2024efficient,
          title={Efficient Test-Time Adaptation of Vision-Language Models},
          author={Karmanov, Adilbek and Guan, Dayan and Lu, Shijian and El Saddik, Abdulmotaleb and Xing, Eric},
          journal={The IEEE/CVF Conference on Computer Vision and Pattern Recognition},
          year={2024}
  }

@article{loshchilov2017decoupled,
  title={Decoupled weight decay regularization},
  author={Loshchilov, Ilya and Hutter, Frank},
  journal={arXiv preprint arXiv:1711.05101},
  year={2017}
}

@inproceedings{zhao2025attention,
  title={Attention Bootstrapping for Multi-Modal Test-Time Adaptation},
  author={Zhao, Yusheng and Luo, Junyu and Luo, Xiao and Huang, Jinsheng and Yuan, Jingyang and Xiao, Zhiping and Zhang, Ming},
  booktitle={Proceedings of the AAAI Conference on Artificial Intelligence},
  pages={22849--22857},
  year={2025}
}

@inproceedings{cherti2023reproducible,
  title={Reproducible scaling laws for contrastive language-image learning},
  author={Cherti, Mehdi and Beaumont, Romain and Wightman, Ross and Wortsman, Mitchell and Ilharco, Gabriel and Gordon, Cade and Schuhmann, Christoph and Schmidt, Ludwig and Jitsev, Jenia},
  booktitle={Proceedings of the IEEE/CVF conference on computer vision and pattern recognition},
  pages={2818--2829},
  year={2023}
}

@article{zhang2022memo,
  title={Memo: Test time robustness via adaptation and augmentation},
  author={Zhang, Marvin and Levine, Sergey and Finn, Chelsea},
  journal={Advances in neural information processing systems},
  volume={35},
  pages={38629--38642},
  year={2022}
}

@article{hendrycks2019benchmarking,
  title={Benchmarking neural network robustness to common corruptions and perturbations},
  author={Hendrycks, Dan and Dietterich, Thomas},
  journal={arXiv preprint arXiv:1903.12261},
  year={2019}
}
}

\clearpage
\appendix
\setcounter{page}{1}
\maketitlesupplementary

\begin{figure*}
    
    \centering
    \includegraphics[width=1\linewidth]{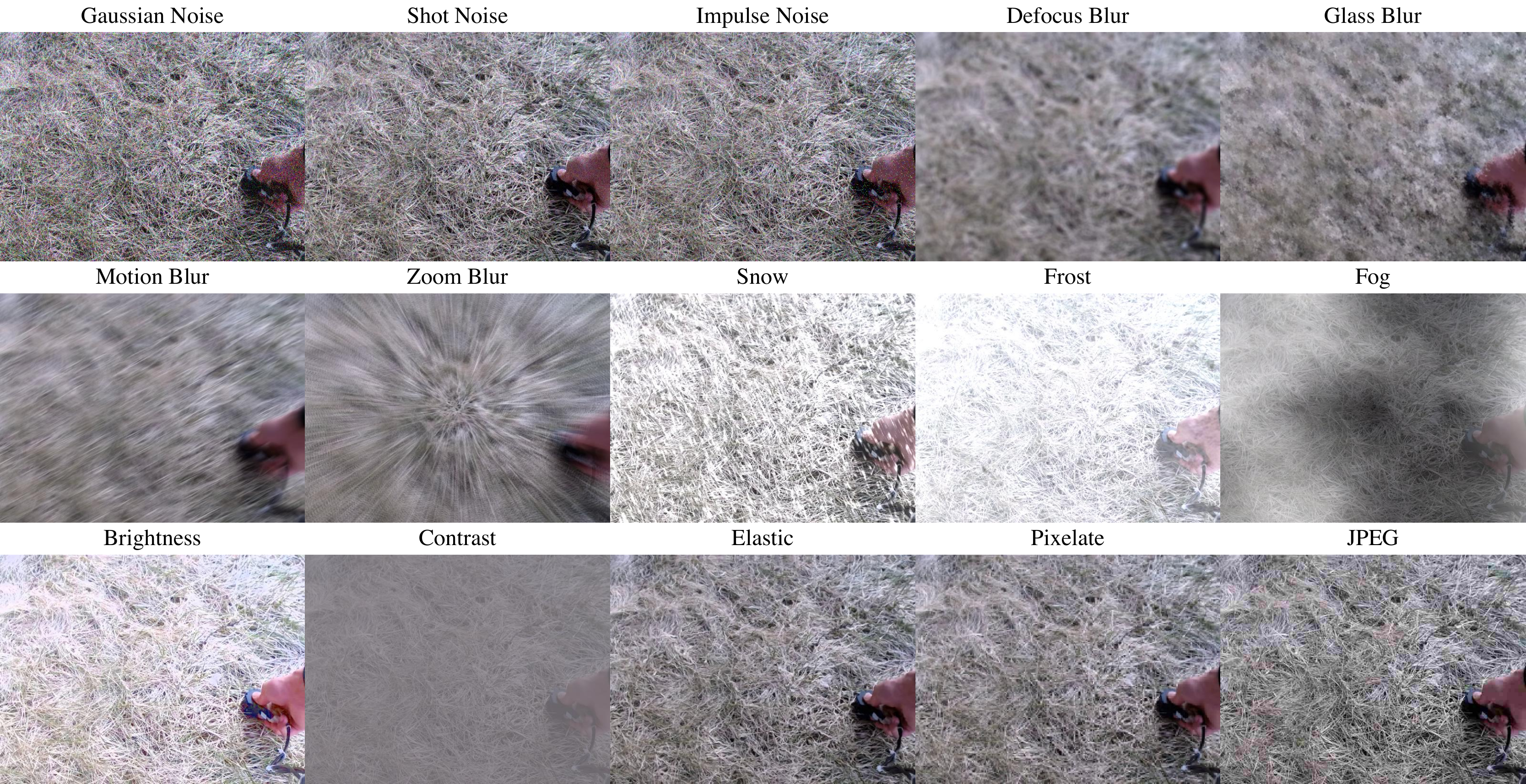}
    \caption{Visualization of the 15 visual corruption types on the \textbf{TAG-C} benchmark.}
    \label{fig:benchmark_c_v}
\end{figure*}

\section{Wild Real-world Corruptions}
In real-world robotics applications, Tactile-Vision-Language (TVL) models often encounter unpredictable environmental factors that introduce asynchronous distribution shifts across modalities. These shifts can manifest as sensor-specific corruptions, such as visual distortions from varying lighting conditions or camera malfunctions, and tactile degradations from physical wear, inconsistent contact pressure, or surface contaminants. To rigorously evaluate test-time adaptation methods under such conditions, our TAG-C benchmark incorporates a diverse set of corruption types designed to mimic these wild, real-world scenarios. Drawing from practical robotic deployments—like disaster response missions where robots navigate foggy or dusty environments, or manipulation tasks involving worn tactile sensors—these corruptions simulate the modality-specific uncertainties that challenge model robustness.

Figure~\ref{fig:benchmark_c_v} illustrates the 15 visual corruption types applied to the TAG-C benchmark, including Gaussian Noise (simulating sensor interference in low-light conditions), Defocus Blur (emulating camera focus issues during rapid movements), Fog (representing atmospheric haze in outdoor settings), and others like Motion Blur (from robot vibrations) and Snow (for adverse weather). These corruptions reflect common real-world visual degradations, such as those encountered in autonomous navigation or object recognition tasks under variable environmental conditions.

\begin{figure}
    \centering
    \includegraphics[width=1\linewidth]{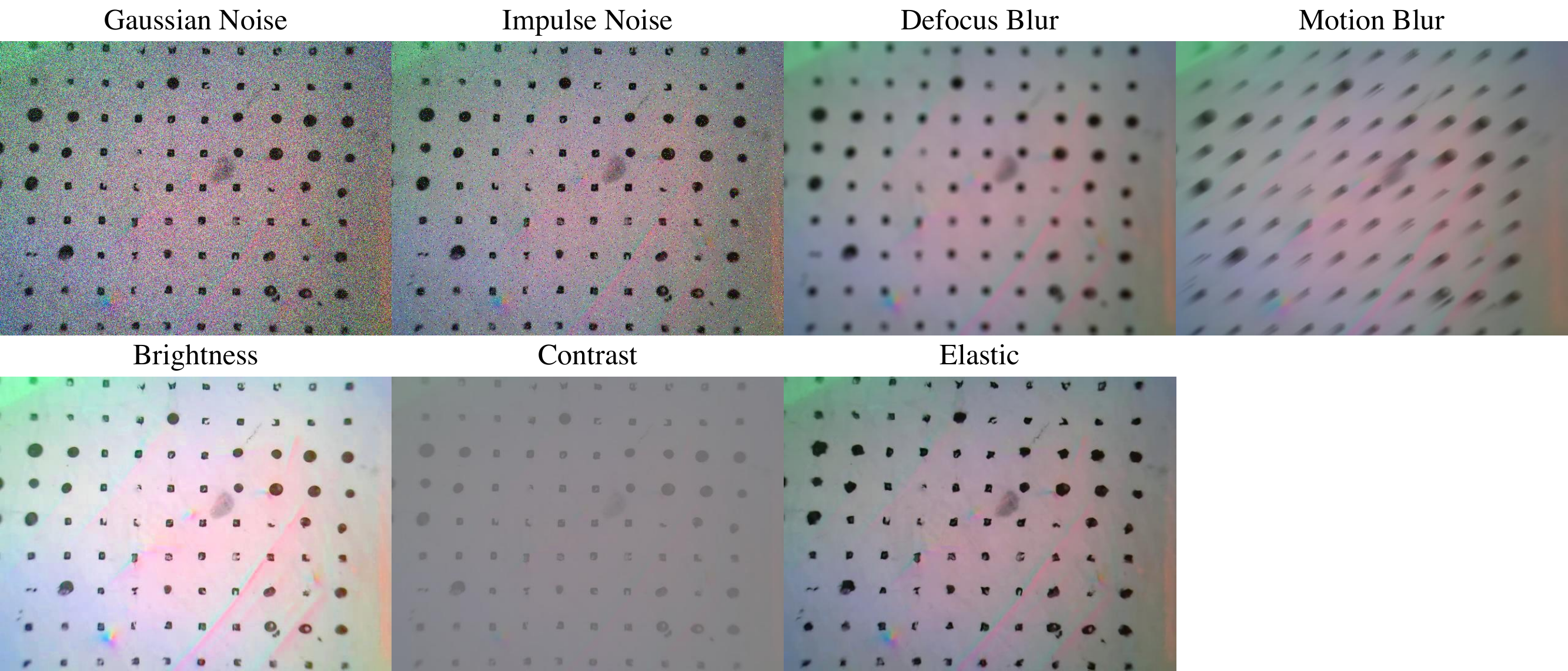}
    \caption{Visualization of the 7 tactile corruption types on the \textbf{TAG-C} benchmark.  }
    \label{fig:benchmark_c_g}
\end{figure}

Figure~\ref{fig:benchmark_c_g} depicts the 7 tactile corruption types, encompassing Brightness (altered sensor response due to lighting on optical-based tactile sensors), Contrast (variations from inconsistent contact force), Elastic (deformations mimicking sensor membrane wear), and more. These emulate practical issues like tactile signal distortion from prolonged use in industrial grasping or exploration in unstructured terrains, where sensors may accumulate dirt or experience mechanical fatigue.

By incorporating these realistic corruptions, TAG-C provides a comprehensive testbed for assessing methods like RobustTouch, ensuring they can adapt to the asynchronous and modality-independent shifts prevalent in wild, unconstrained environments. This benchmark highlights the necessity of reliability-aware adaptation strategies to maintain performance amid such challenges.

\section{Benchmark Details}
The Touch and Go (TAG) dataset \cite{yang2022touch} represents a large-scale, human-collected repository of visuo-tactile data, specifically engineered to advance multi-modal learning by bridging visual and tactile modalities. In contrast to prior datasets, which are often restricted to controlled laboratory environments or narrow robotic manipulation tasks, TAG encompasses paired visual and tactile observations gathered from a wide array of real-world settings via an innovative human-centric collection methodology employing the GelSight sensor.

To simulate realistic asynchronous distribution shifts in Tactile-Vision-Language (TVL) models, we curated 4,126 paired visual-tactile samples from the TAG test set, introduced targeted corruptions, and thereby developed the \textbf{TAG-C} benchmark. For visual corruptions, we adopted the framework from ImageNet-C \cite{hendrycks2019benchmarking}, applying 15 distinct types to the visual samples: Gaussian Noise, Shot Noise, Impulse Noise, Defocus Blur, Glass Blur, Motion Blur, Zoom Blur, Snow, Frost, Fog, Brightness, Contrast, Elastic, Pixelate, and JPEG. For tactile corruptions, we tailored the selection to the inherent properties of the GelSight sensor \cite{yuan2017gelsight}, incorporating only 7 relevant types: Gaussian Noise, Impulse Noise, Defocus Blur, Motion Blur, Brightness, Contrast, and Elastic. Furthermore, each modality-specific corruption is implemented across 5 severity levels to capture a broader spectrum of real-world variabilities. Figures \ref{fig:benchmark_c_v} and \ref{fig:benchmark_c_g} offer a detailed visualization of the TAG-C benchmark, illustrating the corruption types for the visual and tactile modalities, respectively.
\section{Pseudocode}

Algorithm~\ref{alg:robusttouch} provides the pseudocode for the RobustTouch method, detailing the core procedures for dynamic reliable sample filtering, quality-aware dynamic modality fusion, and reliability-aware loss computation. This structured overview illustrates how the algorithm processes each test mini-batch, computes reliability indicators, performs adaptive fusion, and updates the model parameters during test-time adaptation in TVL models under asynchronous distribution shifts.

\begin{algorithm}[h]
\caption{RobustTouch Algorithm}
\label{alg:robusttouch}
\begin{algorithmic}[1]
\STATE Initialize model $\Theta$, fusion network $G(\cdot)$, optimizers, hyperparameters ($\alpha$, $\lambda$, $\gamma$, $N$)
\FOR{each test mini-batch ${x_v, x_t}$}
\STATE Compute clean embeddings $e_v, e_t$ and affinities $a_v, a_t$
\STATE Perturb inputs: $x_v' \gets \textsc{Perturb}(x_v)$, $x_t' \gets \textsc{Perturb}(x_t)$
\STATE Compute perturbed embeddings $e_v', e_t'$ and affinities $a_v', a_t'$ (no gradients)
\FOR{$m \in {v, t}$}
\STATE $U_m \gets -\sum p_i \log p_i$ where $p_i = \delta(a_m)$
\STATE $V_m \gets \max(\delta(a_m)) - \max(\delta(a_m'))$
\ENDFOR
\STATE Form $\nabla \gets [V_t, U_t, V_v, U_v]$
\FOR{$m \in {v, t}$}
\STATE Thresholds: $\zeta^U_m \gets \mu^U_m + \alpha \sigma^U_m$, $\zeta^V_m \gets \mu^V_m - \alpha \sigma^V_m$
\STATE Mask: $M^i_m \gets 1$ if $U_m < \zeta^U_m$ and $V_m > \zeta^V_m$ else $0$
\ENDFOR
\STATE Sample mask: $M^i \gets \prod_{m} M^i_m$
\STATE Detach unreliable samples ($M^i = 0$)
\STATE Fusion weights: $[w_v, w_t] \gets G(\nabla)$
\STATE Fused embedding: $e_{fus} \gets w_v \frac{e_v}{|e_v|2} + w_t \frac{e_t}{|e_t|2}$
\STATE Fused logits: $z{fus} \gets A(e{fus}, \frac{e_l}{|e_l|2})$
\STATE Reliable set: $S \gets {i \mid M^i = 1}$
\FOR{$m \in {v, t}$}
\STATE $L{CR}(a_m) \gets \frac{1}{|S|} \sum_{i \in S} c^i \log \left( \frac{e^\gamma}{c^i} \right)$
\STATE $L_{CB}(a_m) \gets \hat{p} \cdot \log \hat{p}$ where $\hat{p} = \delta \left( \sum_{i \in S} \delta(a^i_m) \right)$
\ENDFOR
\STATE $L_{modal} \gets \sum_m (L_{CR}(a_m) + \lambda L_{CB}(a_m))$
\STATE Update branches with $L_{modal}$
\STATE Accumulate $L_{fus} \gets L_{CR}(z_{fus}) + \lambda L_{CB}(z_{fus})$
\IF{batch index mod $N == 0$}
\STATE Update fusion with accumulated $L_{fus}$
\ENDIF
\ENDFOR
\STATE \textbf{return} adapted model
\end{algorithmic}
\end{algorithm}

\section{Learning Rate Details}

To assess the robustness of RobustTouch to variations in learning rate, we conducted sensitivity experiments on the TAG Vision Corruption and TAG Tactile Corruption datasets. We evaluated performance across three learning rates: $1 \times 10^{-6}$, $1 \times 10^{-9}$, and $1 \times 10^{-12}$. These experiments compare RobustTouch against baseline methods including No Adaptation (source), TENT, SAR, TDA, and READ under the continuous cross-domain setting with corrupted modalities (visual for TAG Vision Corruption and tactile for TAG Tactile Corruption).

The results, presented in Table~\ref{tab:lr}, demonstrate that RobustTouch maintains or improves performance as the learning rate decreases, achieving higher accuracy at lower rates (e.g., 63.8\% on TAG Vision Corruption at $1 \times 10^{-12}$). This suggests our method's adaptation mechanisms are less sensitive to aggressive gradient updates, potentially due to the reliability-aware loss and dynamic filtering that stabilize updates even with smaller steps. In contrast, methods like TENT and READ degrade significantly at lower rates, often approaching No Adaptation performance, highlighting their reliance on larger learning rates for effective adaptation. These findings underscore RobustTouch's suitability for real-world TVL applications where computational constraints may necessitate conservative learning rates.

\begin{table}[h]
\centering
\caption{Sensitivity experiment on datasets (TAG Tactile Corruption and TAG Vision Corruption) with respect to learning rates.}
\begin{tabularx}{0.5\textwidth}{lXXXXXX}

\toprule
\textbf{LR} & \textbf{Source} & \textbf{TENT} & \textbf{SAR} & \textbf{TDA} & \textbf{READ} & \textbf{Ours} \\
\midrule
\multicolumn{7}{c}{\textbf{TAG Vision Corruption}} \\
\midrule
\textbf{1e-6} & 57.85 & 23.70 & 57.83 & 32.08 & 11.82 & \textbf{61.68} \\
\textbf{1e-9} & 57.85 & 57.79 & 57.80 & 32.08 & 57.85 & \textbf{63.31} \\
\textbf{1e-12} & 57.85 & 57.81 & 57.80 & 32.08 & 57.82 & \textbf{63.80} \\
\midrule
\multicolumn{7}{c}{\textbf{TAG Tactile Corruption}} \\
\midrule
\textbf{1e-6} & 60.69 & 24.74 & 60.62 & 30.11 & 16.07 & \textbf{67.59} \\
\textbf{1e-9} & 60.69 & 60.67 & 60.61 & 30.11 & 60.70 & \textbf{68.32} \\
\textbf{1e-12} & 60.69 & 60.62 & 60.61 & 30.11 & 60.64 & \textbf{68.56} \\
\bottomrule
\end{tabularx}
\label{tab:lr}
\end{table}


\end{document}